\documentclass[11pt]{article}

\usepackage[preprint]{acl}

\usepackage{times}
\usepackage{latexsym}

\usepackage[T1]{fontenc}

\usepackage[utf8]{inputenc}

\usepackage{microtype}

\usepackage{inconsolata}

\usepackage{graphicx}

\usepackage{graphicx}
\usepackage{amsmath}
\usepackage{amssymb}
\usepackage{mathtools}
\usepackage{amsthm}
\usepackage{bbm}
\usepackage{algorithm}
\usepackage{algorithmic}  
\usepackage{enumitem}
\usepackage{multirow}
\usepackage{booktabs}
\usepackage[dvipsnames]{xcolor}
\usepackage[most]{tcolorbox}
\usepackage{fontawesome}
\usepackage[normalem]{ulem}
\useunder{\uline}{\ul}{}
\usepackage{subcaption}
\usepackage[table,xcdraw]{xcolor}
\usepackage{xurl}
\usepackage{hyperref}

\newcommand{\model}{BGPO}
\newcommand{\expectation}{\mathbb{E}_{x \sim \mathcal{D}, y \sim \pi_{\theta_{\text{old}}}(\cdot \mid x)}}
\newtheorem{lemma}{Lemma}

\title{Boundary-Guided Policy Optimization for Memory-efficient RL of Diffusion Large Language Models}

\author{
  Nianyi Lin\thanks{Equal contribution.~Work done when JZ interned at Zhipu.},
  Jiajie Zhang\footnotemark[1],
  Lei Hou,
  Juanzi Li \\
  Tsinghua University
}

\begin{document}
\maketitle

\begin{abstract}
A key challenge in applying reinforcement learning (RL) to diffusion large language models (dLLMs) is the intractability of their likelihood functions, which are essential for the RL objective, necessitating corresponding approximation during training.
While existing methods approximate the log-likelihoods by their evidence lower bounds (ELBOs) via customized Monte Carlo (MC) sampling, they incur significant memory overhead due to the need to retain all MC samples for the gradient computation of non-linear terms in the RL objective, and thus restrict feasible sample sizes, leading to imprecise likelihood approximations and distorted RL objective. 
To address this, we propose \emph{Boundary-Guided Policy Optimization} (BGPO), a memory-efficient RL algorithm that maximizes a specially constructed lower bound of the ELBO-based objective. This lower bound is carefully designed to satisfy two key properties: (1) Linearity: it is a linear sum where each term depends only on a single MC sample, thereby enabling gradient accumulation across samples and ensuring constant memory usage; (2) Equivalence: Both the value and gradient of this lower bound are equal to those of the ELBO-based objective in on-policy training, making it also an effective approximation for the original RL objective. These properties allow BGPO to adopt a large MC sample size, improving likelihood approximations and RL objective estimation, which in turn leads to enhanced performance. Experiments show that BGPO significantly outperforms previous RL algorithms for dLLMs in math problem solving, code generation, and planning tasks. 
Our codes and models are available at
\href{https://github.com/THU-KEG/BGPO}{https://github.com/THU-KEG/BGPO}.
\end{abstract}

\section{Introduction}

\begin{figure*}[!t]
    \centering
    \raisebox{-2pt}{
    \begin{subfigure}{0.33\linewidth}
        \centering
        \includegraphics[width=\linewidth]{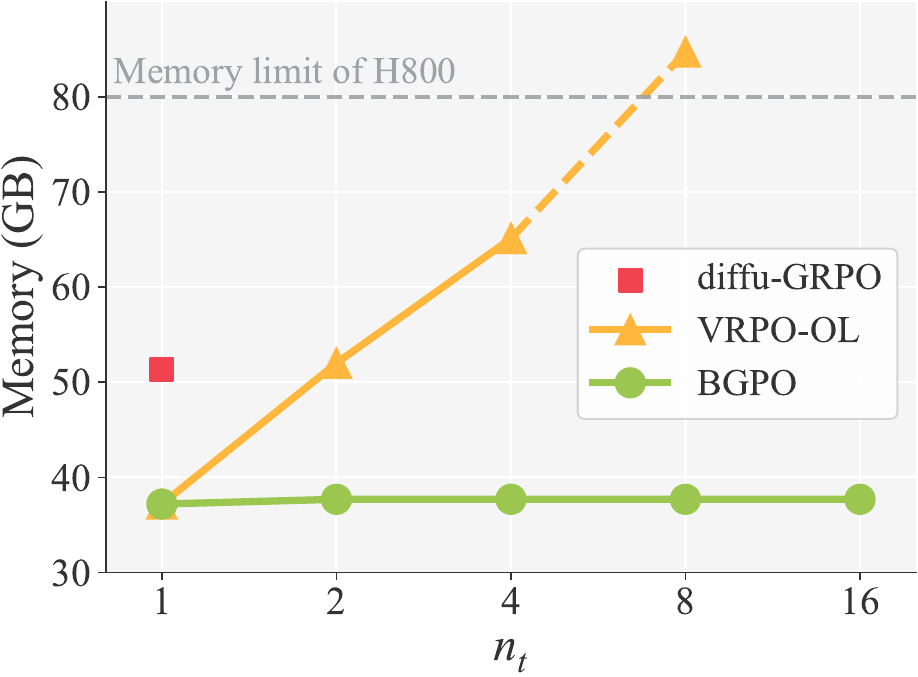}
    \end{subfigure}
    }
    \begin{subfigure}{0.63\linewidth}
        \centering
        \includegraphics[width=\linewidth]{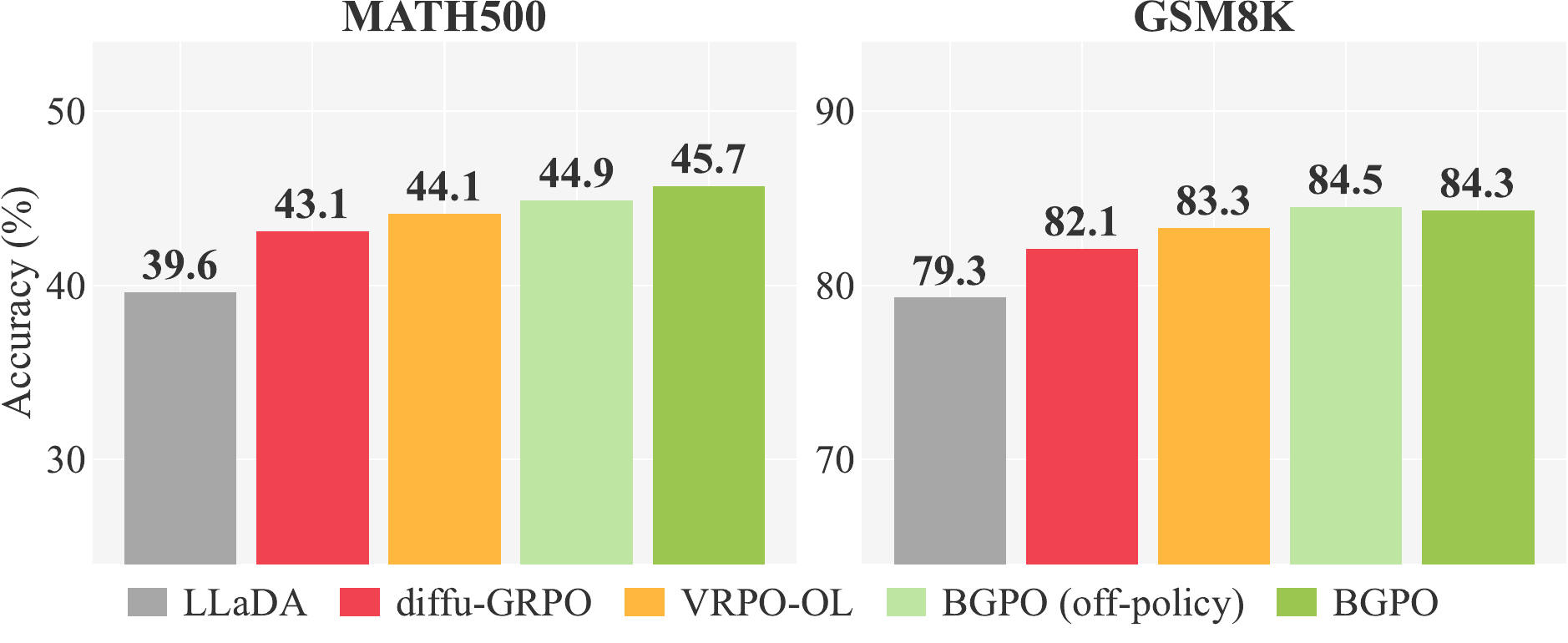}
    \end{subfigure}
    \caption{Left: Comparison of memory usage of previous ELBO-based RL method (VRPO-OL) and our BGPO using different Monte Carlo sample size $n_t$ for the RL objective approximation. The max response length is set to 512. Middle and right: Performance of LLaDAs with different RL algorithms on mathematical tasks.}
    \vspace{-0.3cm}
    \label{fig:intro}
\end{figure*}

Recently, diffusion large language models (dLLMs) have emerged as promising alternatives to conventional autoregressive models (ARMs), demonstrating competitive performance in various language modeling tasks~\cite{llada1.0, dream7b, diffucoder, sdar}. Unlike ARMs, which generate sequences in a left-to-right, token-by-token manner, dLLMs iteratively unmask tokens in parallel, offering the potential for significant inference acceleration~\cite{deepmind2025gemini, inception2025mercury, seed-diffusion, fast-dllm}. Despite these advancements, existing work focuses mainly on pre-training and supervised fine-tuning of dLLMs, while leveraging reinforcement learning (RL) to further enhance dLLMs remains a challenging problem, even though RL has demonstrated significant efficacy in improving various capabilities of LLMs~\cite{openai2024reasoning, deepseek-r1}.

A key challenge in applying RL to dLLMs is the intractability of their likelihood functions~\cite{llada1.5, d1, wd1}. Specifically, the iterative, non-sequential generation process precludes exact calculation of the log-likelihood for generated responses~\cite{llada1.5, d1, wd1}, which is essential for RL algorithms~\cite{ppo, grpo}. In light of this, recent work has explored approximating log-likelihoods by their evidence lower bounds (ELBOs) via customized Monte Carlo (MC) sampling~\cite{llada1.5}. While increasing the MC sample size can yield highly accurate approximations~\cite{ho2020, song2025score}, this approach incurs substantial memory overhead during RL training, as it requires storing the forward computational graphs for all MC samples to compute the gradient of non-linear terms in the RL objective. As a result, practical implementations can only adopt relatively small sample sizes (e.g., $n_t=4$ as illustrated in the left of Figure~\ref{fig:intro}) due to hardware constraints, which directly amplifies errors in log-likelihood approximation and introduces substantial bias and variance for the estimated objective and its gradients, ultimately degrading performance. 

To address this limitation, we propose \emph{Boundary-Guided Policy Optimization} (BGPO), a memory-efficient RL algorithm for dLLMs that supports large MC sample sizes for log-likelihood and RL objective approximation. Specifically, BGPO maximizes a constructed lower bound of the ELBO-based objective. This lower bound is carefully designed to satisfy two critical properties: (1) Linearity: it is formulated in a linear sum where each term associates with a single MC sample, thereby enabling gradient accumulation across samples and ensuring constant memory usage irrespective of sample size; (2) Equivalence: Both the value and gradient of the lower bound are equal to those of the ELBO-based objective in on-policy training, ensuring that the lower bound can also effectively approximate the original RL objective. These properties allow BGPO to adopt a large MC sample size to obtain a more accurate approximation for the RL objective, thereby achieving better performance.

To validate the effectiveness of BGPO, we conduct RL experiments with LLaDA-8B-Instruct on math problem solving, code generation, and planning tasks. The results show that BGPO significantly improves the performance of LLaDA-8B-Instruct across all tasks and also outperforms previous RL algorithms for dLLMs. Further analysis demonstrates that increasing the MC sample size effectively reduces the bias and variance of gradients and improves model performance. 
Notably, BGPO achieves these improvements with only marginal increases in average training step time, despite its larger sample size.

In summary, our main contributions include: 
(1) We propose BGPO, a memory-efficient RL algorithm for dLLMs that supports large MC sample sizes in the approximation of log-likelihoods and the RL objective;
(2) We theoretically prove the equivalence of the BGPO objective and the ELBO-based objective in on-policy training, demonstrating that BGPO also provides an effective approximation of the original RL objective; 
(3) Through comprehensive experiments, we validate the efficacy of BGPO and demonstrate the value of larger MC sample sizes in boosting model performance.
We hope our work establishes a firm foundation for future research on RL for dLLMs.

\vspace{-0.1cm}
\section{Preliminary}
\label{sec:preliminary}
\subsection{Masked Diffusion Language Models}
Masked dLLMs employ a non-autoregressive generation paradigm, generating text through progressive denoising. At their core lies a mask predictor $p_\theta$ \cite{austin2023structured, ou2024your}, which learns the data distribution via a forward-reverse framework. Starting from the original text at $t=0$, the forward process gradually masks the input tokens until the sequence is fully masked at $t=1$. Following LLaDA~\cite{llada1.0}, at time $t\in (0, 1)$, each token is replaced by the mask token $\mathbf{M}$ with probability $t$ and remains unmasked with probability $1-t$. In the reverse process, the mask predictor recovers this sequence by iteratively predicting the masked tokens as time reverses from 1 to 0. In conditional generation scenarios, the prompt $x$ always remains unmasked, and the forward-reverse process is only applied to the response $y$. 

\subsection{Challenges of Applying RL to dLLMs.}
Reinforcement learning (RL) has proved effective for improving language models. The basic objective is to maximize: 
\begin{align}
\mathcal{J}(\theta) = &\mathbb{E}_{x \sim \mathcal{D}, y \sim \pi_{\theta}(\cdot | x)}A(x, y) \notag \\
= &\expectation{\textstyle\frac{\pi_\theta(y|x)}{\pi_{\theta_\text{old}}(y|x)}}A(x, y), \notag \\
= &\expectation \mathcal{R}(x,y), 
\label{eq:rl_obj}
\end{align}
where $\pi_\theta$, $\pi_{\theta_\text{old}}$, and $A(x,y)$ denote the current policy, old policy, and sequence-level advantage, respectively, 
and
\begin{align}
\mathcal{R}(x,y) = e^{\log\pi_\theta(y|x)-\log\pi_{\theta_\text{old}}(y|x)}A(x, y).
\label{eq:policy_ratio}
\end{align}
However, applying RL to dLLMs is nontrivial, since the iterative denoising generation makes the exact computation of $\log\pi_\theta(y|x)$ intractable. 

To address this, recent work has developed several methods to approximate $\log\pi_\theta(y|x)$. diffu-GRPO~\cite{d1} adopts a single-pass estimation, simply making $\log\pi_\theta(y|x)=\sum_{i=1}^{|y|}\log p_\theta(y^i|x')$, where $y^i$ is the $i$-th token of $y$ and $x'$ is a randomly masked prompt. Though efficient, it introduces notable bias relative to the exact policy. Alternatively, VRPO~\cite{llada1.5} proposes to approximate $\log\pi_\theta(y|x)$ using its evidence lower bound (ELBO): 
\begin{equation}
    B_{\pi_\theta}(y | x) \triangleq \mathbb{E}_{t \sim \mathcal{U}[0,1], y_t \sim q(\cdot | t, y, x)} \ell_{\pi_\theta}(y_t, t, y | x),
\label{eq:elbo}
\end{equation}
where $q(\cdot | t, y, x)$ denotes the forward masking process for the response $y$ at time $t$, and
\begin{equation}
\label{eq:per_step_loss}
    \ell_{\pi_\theta}(y_t, t, y | x) \triangleq \frac{1}{t} \sum_{i=1}^{|y_t|} \mathbf{1}[y_t^i \!=\! \mathbf{M}] \log p_\theta(y^i | y_t, x).
\end{equation}
Specifically, they estimate $B_\pi(y|x)$ via customized Monte Carlo sampling:
\begin{equation}
\hat{B}_{\pi_\theta}(y | x) \!=\! \frac{1}{n_t} \sum_{j=1}^{n_t} \ell_{\pi_\theta}(y_{t^{(j)}}, t^{(j)}, y | x),
\label{eq:mcs_elbo}
\end{equation}
where $t^{(j)}\overset{\mathrm{i.i.d.}}{\sim} \mathcal{U}[0,1]$ and $y_{t^{(j)}} \overset{\mathrm{i.i.d.}}{\sim} q(\cdot | t^{(j)}, y, x)$ are the sampled timestamps and corresponding partially masked responses.
Substituting $\hat{B}_{\pi_\theta}(y | x)$ into Eq.~\ref{eq:rl_obj} yields an approximated RL objective:
\begin{align}
\hat{\mathcal{J}}(\theta) = \expectation \hat{\mathcal{R}}(x,y), 
\label{eq:vrpo_obj}
\end{align}
where
\begin{align} 
\hat{\mathcal{R}}(x,y) = e^{\hat{B}_{\pi_\theta}(y | x) - \hat{B}_{\pi_{\text{old}}}(y | x)} A(x, y).
\label{eq:elbo_ratio}
\end{align}
Notably, previous work has shown that when the sample size $n_t$ is large enough,  the bias of $\hat{B}_{\pi}(y|x)$ for a well-trained model relative to $\log\pi_\theta(y|x)$ will become negligible~\cite{ho2020, song2025score}. However, using a large $n_t$ in training requires a huge amount of GPU memory: Each time $\hat{\mathcal{R}}(x,y)$ is computed, $n_t$ forward passes of $p_\theta$ need to be executed (i.e., Eq.~\ref{eq:per_step_loss} and~\ref{eq:mcs_elbo}), and all the $n_t$ computational graphs must be retained in memory to calculate the gradient of the exponential function in Eq.~\ref{eq:elbo_ratio}. As a result, in practice, the sample size can only remain small (e.g., $n_t=4$), which results in inaccurate approximations for the likelihoods as well as the final objective, seriously affecting the final performance.

To break through this limitation, we propose \emph{Boundary-Guided Policy Optimization} (BGPO), a memory-efficient RL algorithm for dLLMs that supports a large Monte Carlo sample size, thereby enabling more accurate approximations and achieving better performance. A detailed introduction is provided in the following sections.

\section{\model}
\label{sec:method}
Following~\citet{llada1.5}, our BGPO algorithm also uses the estimated ELBO $\hat{B}_{\pi_\theta}(y | x)$ to approximate $\log \pi_\theta(y|x)$.  The main difference is that instead of directly maximizing the approximated objective $\hat{\mathcal{J}}(\theta)$, BGPO turns to maximize a constructed tight lower bound of $\hat{\mathcal{J}}(\theta)$:
\begin{align}
\hat{\mathcal{J}}_\text{lb}(\theta) = \expectation \hat{\mathcal{R}}_\text{lb}(x,y), 
\label{eq:bgpo_obj}
\end{align}
where $\hat{\mathcal{R}}_\text{lb}(x,y) \!\leq\! \hat{\mathcal{R}}(x,y)$. Specifically, $\hat{\mathcal{R}}_\text{lb}(x,y)$ is carefully designed so that it satisfies the following two properties\footnote{For simplicity, we mainly discuss $\hat{\mathcal{R}}_\text{lb}$ and $\hat{\mathcal{R}}$ in this section, while all their properties can directly apply to $\hat{\mathcal{J}}_\text{lb}$ and $\hat{\mathcal{J}}$, unaffected by the expectation function.}:
\begin{itemize}[itemsep=0pt, leftmargin=*]
\item{Linearity:} $\hat{\mathcal{R}}_\text{lb}(x,y)$ is formulated as $\sum_{j=1}^{n_t} g_j$, where $g_j$ is a function of the partially masked sample $y_{t^{(j)}}$ at time $t^{(j)}$. Therefore, we can backpropagate the gradient of $g_j$ for each $y_{t^{(j)}}$ separately and update the policy after all backward passes, so that the memory usage becomes irrelevant to the MC sample size $n_t$.
\item{Equivalence:} In on-policy training (i.e., $\pi_{\theta_\text{old}}=\pi_{\theta}$), the value and gradient of $\hat{\mathcal{R}}_\text{lb}(x,y)$ are always equal to those of $\hat{\mathcal{R}}(x,y)$, making $\hat{\mathcal{J}}_\text{lb}(\theta)$ equivalent to $\hat{\mathcal{J}}(\theta)$ and also an effective approximation of the original RL objective $\mathcal{J}(\theta)$. 
\end{itemize}
These two properties allow BGPO to use a larger MC sample size in the likelihood approximation, which effectively reduces the bias and variance of $\hat{\mathcal{J}}_\text{lb}(\theta)$ and its gradient, leading to better performance. In the following, we will introduce the construction of $\hat{\mathcal{R}}_\text{lb}(x,y)$ in detail.

\subsection{Linear Lower Bound Construction}
The construction of $\hat{\mathcal{R}}_\text{lb}(x,y)$ is different based on the sign of the advantage $A(x,y)$: 
\begin{itemize}[itemsep=0pt, leftmargin=*]
\item For $A(x,y) \ge 0$, we construct $\hat{\mathcal{R}}_\text{lb}(x,y)$ using Taylor expansion;
\item For $A(x,y) < 0$, we construct $\hat{\mathcal{R}}_\text{lb}(x,y)$ using Jensen's inequality.
\end{itemize}

\begin{lemma}\textbf{[First-order Taylor Expansion]}
\label{lemma:taylor}
For any $\delta \in \mathbb{R}$, the exponential function satisfies
\[e^{\delta} \geq 1 + \delta.\]
\end{lemma}

\noindent When $A(x, y) \ge 0$, we apply the first-order Taylor expansion in Eq.~\ref{eq:elbo_ratio}, which yields:
\begin{align} 
\hat{\mathcal{R}}(x,y) & = e^{\hat{B}_{\pi_\theta}(y | x) - \hat{B}_{\pi_{\text{old}}}(y | x)} A(x, y) \notag \\
& = e^{\left(\frac{1}{n_t}\sum_{j=1}^{n_t}d_j\right)} A(x, y) \notag \\
& \ge \left(1 + \frac{1}{n_t}\sum\limits_{j=1}^{n_t}d_j\right) A(x,y) \notag \\
& = \sum_{j=1}^{n_t}\frac{(1+d_j)A(x,y)}{n_t},
\label{eq:apply_taylor}
\end{align}
where
\begin{align}
d_j = \ell_{\pi_\theta}(y_{t^{(j)}}, t^{(j)}, y | x) -  \ell_{\pi_{\theta_\text{old}}}(y_{t^{(j)}}, t^{(j)}, y | x).
\end{align}

\begin{lemma}\textbf{[Jensen's Inequality]}
\label{lemma:jensen}
For a convex function $f$ and a finite set $\{x_i\}_{i=1}^n$, we have 
\begin{equation}
    \label{eq:jensen}
    f\left(\frac{1}{n}\sum_{i=1}^n x_i\right) \leq \frac{1}{n}\sum_{i=1}^n f(x_i).
\end{equation}
\end{lemma}

\noindent When $A(x, y) < 0$, by applying Jensen's inequality in Eq.~\ref{eq:elbo_ratio}, we have:
\begin{align} 
\hat{\mathcal{R}}(x,y) & = e^{\left(\frac{1}{n_t}\sum_{j=1}^{n_t}d_j\right)} A(x, y) \notag \\
& \ge \left(\frac{1}{n_t}\sum_{j=1}^{n_t}e^{d_j}\right) A(x,y) \notag \\
& = \sum_{j=1}^{n_t}\frac{e^{d_j}A(x,y)}{n_t}.
\label{eq:apply_jensen}
\end{align}
Putting everything together and letting:
\begin{equation}
    g_j \!=\! 
    \left\{
    \begin{aligned}
      &\scalebox{1.1}{$\frac{(1+d_j)A(x,y)}{n_t}$}, & \text{if } A(x,y) \ge 0; \\
      &\scalebox{1.1}{$\frac{e^{d_j}A(x,y)}{n_t}$}, & \text{if } A(x, y) < 0,
    \end{aligned}
    \right.
\label{eq:g}
\end{equation}
$\hat{\mathcal{R}}_\text{lb}(x,y)$ is constructed as a linear sum of $g_j$:
\begin{equation}
    \hat{\mathcal{R}}_\text{lb}(x,y) = \sum_{j=1}^{n_t} g_j.
\label{eq:lower_bound_def}
\end{equation}
As shown in Algorithm~\ref{alg:BGPO}, the linearity of $\hat{\mathcal{R}}_\text{lb}(x,y)$ (as well as $\hat{\mathcal{J}}_\text{lb}(\theta)$) enables us to separate the gradient backpropagation for each $y_{t^{(j)}}$, thus keeping the memory usage constant and allowing a larger sample size $n_t$.

\begin{algorithm*}[!t]
\caption{\textit{\model}}
\small
\label{alg:BGPO}
\textbf{Input:} dataset $\mathcal{D}$; initial policy model $\pi_\theta$; 
hyperparameters: $G$, $n_t$, $\eta$. 
\begin{algorithmic}[1]
    \FOR{iteration $=1,2,\dots, M$}
        \STATE Update the old policy $\pi_{\theta_\text{old}} \gets \pi_\theta$
        and sample a batch $\mathcal{D}_b$ from $\mathcal{D}$
        \FOR{each prompt $x\in\mathcal{D}_b$}
            \STATE Sample $G$ response $\{y^{(i)}\}_{i=1}^{G} \sim \pi_{\theta_{\text{old}}}(\cdot | x^{(b)})$ and compute advantages $\{A(x, y^{(i)})\}_{i=1}^{G}$ using Eq.~\ref{eq:adv}
            \FOR{ $i=1$ \TO $G$}
                \STATE Sample $n_t$ timestamp $\{t^{(j)}\}_{j=1}^{n_t} \sim U[0, 1]$ 
                \FOR{$j = 1$ \TO $n_t$}
                    \STATE Sample partially masked response $y^{(i)}_{t^{(j)}} \sim q(\cdot | t^{(j)}, y^{(i)}, x)$
                    \STATE Compute $g_j$ using Eq.~\ref{eq:g} and let $\mathcal{L}_{j} \gets -\frac{g_j}{G}$
                    \STATE Backpropagate the gradient  of $\mathcal{L}_{j}$ \ \ ($\triangleright$ graident accumulation)
                \ENDFOR
            \ENDFOR
        \ENDFOR
        \STATE Update the policy $\theta \gets \theta - \eta \nabla_\theta$
    \ENDFOR
\end{algorithmic}
\textbf{Output:} $\pi_\theta$ 
\end{algorithm*}

\subsection{Proof of Equivalence}
In on-policy training where $\pi_{\theta} = \pi_{\theta_\text{old}}$, the value of $\ell_{\pi_{\theta}}$ is equal to $\ell_{\pi_{\theta_\text{old}}}$, which means the value of $d_j$ is always 0. By applying this to Eq.~\ref{eq:elbo_ratio} and~\ref{eq:lower_bound_def}, we can find the values of $\hat{\mathcal{R}}_\text{lb}(x,y)$ and $\hat{\mathcal{R}}(x,y)$ are both equal to $A(x,y)$. 
Moreover, the gradient of $\hat{\mathcal{R}}_\text{lb}(x,y)$ is also the same as that of $\hat{\mathcal{R}}(x,y)$ when $d_j=0$. Specifically, by applying the chain rule of the derivative, we have:
\begin{align}
    \nabla_\theta \hat{\mathcal{R}}(x, y) & = \nabla_\theta \left(e^{\left(\frac{1}{n_t}\sum_{j=1}^{n_t}d_j\right)} A(x, y)\right) \notag \\
    & = e^{\left(\sum_{j=1}^{n_t}\frac{d_j}{n_t}\right)}\frac{A(x, y)\nabla_\theta\left(\sum_{j=1}^{n_t} d_j\right)}{n_t} \notag \\
    & \overset{d_j=0}{=} \sum_{j=1}^{n_t}\frac{A(x,y)\nabla_\theta d_j}{n_t}. 
\end{align}
Similarly, when $A(x, y) \ge 0$, we have:
\begin{align}
    \nabla_\theta \hat{\mathcal{R}}_\text{lb}(x, y) & = \nabla_\theta \left(\sum_{j=1}^{n_t}\frac{(1+d_j)A(x,y)}{n_t}\right) \notag \\
    & = \sum_{j=1}^{n_t}\frac{A(x,y)\nabla_\theta d_j}{n_t},
\end{align}
and when $A(x, y) < 0$, we have:
\begin{align}
    \nabla_\theta \hat{\mathcal{R}}_\text{lb}(x, y) & = \nabla_\theta \left(\sum_{j=1}^{n_t}\frac{e^{d_j}A(x,y)}{n_t}\right) \notag \\
    & = \sum_{j=1}^{n_t} \frac{A(x, y)e^{d_j}\nabla_\theta d_j}{n_t} \notag \\
    & \overset{d_j=0}{=} \sum_{j=1}^{n_t}\frac{A(x,y)\nabla_\theta d_j}{n_t}. 
\end{align}
Therefore, $\hat{\mathcal{R}}_\text{lb}(x,y)$ and $\hat{\mathcal{R}}(x,y)$ (as well as $\hat{\mathcal{J}}_\text{lb}(\theta)$ and $\hat{\mathcal{J}}(\theta)$) are equivalent in terms of both value and gradient in on-policy training. This means like $\hat{\mathcal{J}}(\theta)$, $\hat{\mathcal{J}}_\text{lb}(\theta)$ is also an effective approximation of $\mathcal{J}(\theta)$, and using a large sample size $n_t$ can reduce the bias and variance of $\hat{\mathcal{J}}_\text{lb}(\theta)$ and its gradient, leading to better model performance.

RL training often adopts \emph{off-policy} optimization to improve sample efficiency. In this setting, where $\pi_{\theta} \neq \pi_{\theta_{\text{old}}}$, though the equivalence between $\hat{\mathcal{R}}_{\text{lb}}(x,y)$ and $\hat{\mathcal{R}}(x,y)$ no longer holds because $d_j \neq 0$, optimizing $\hat{\mathcal{R}}_{\text{lb}}(x,y)$ as a lower bound of $\hat{\mathcal{R}}(x,y)$ still proves effective in driving policy improvement, as shown by the experiments in Section~\ref{sec:experiment}.

\subsection{Final Loss of BGPO}
In practice, we adopt group-based advantage estimation. Specifically, for each prompt $x$, we sample $G$ responses $y^{(1)}, \dots, y^{(G)}$ from $\pi_{\theta_\text{old}}(\cdot | x)$. Let $r(x, y^{(i)})$ denotes the reward of $y^{(i)}$. The advantage of $y^{(i)}$ is defined as:
\begin{equation}
    A(x, y^{(i)}) = \scalebox{1.1}{$\frac{r(x, y^{(i)}) - \text{mean}(\{r(x, y^{(j)})\}_{j=1}^G)}{\text{std}(\{r(x, y^{(j)})\}_{j=1}^G)}$}.
    \label{eq:adv}
\end{equation}
Accordingly, the loss for BGPO is formulated as:
\begin{align}
    \mathcal{L}_\text{BGPO} \!=\! -\mathbb{E}_{\scalebox{0.8}{$\substack{x\sim\mathcal{D},\\\{y^{(i)}\}_{i=1}^G\sim\pi_{\theta_\text{old}}(\cdot | x)}$}}\left[\frac{1}{G} \sum\limits_{i=1}^G \hat{\mathcal{R}}_\text{lb}(x, y^{(i)})\right].
\end{align}
Finally, we summarize our BGPO algorithm in Algorithm~\ref{alg:BGPO}.

\section{Experiment}
\label{sec:experiment}
In this section, we empirically validate the efficacy of BGPO through extensive RL experiments.

\subsection{Setup}

\paragraph{Models.} We employ LLaDA-8B-Instruct~\cite{llada1.0}, a state-of-the-art dLLM that has undergone pre-training and supervised fine-tuning, as our initial policy model.

\paragraph{Datasets.} We conduct RL experiments in three domains: math problem solving, code generation, and planning tasks~\cite{dream7b}. For math problem solving, we train the model on a mix of the training splits of MATH~\cite{math} and GSM8K~\cite{gsm8k}, and evaluate on the respective test sets. For code generation, we use 16K medium-difficulty problems filtered from DeepCoder~\cite{deepcoder2025} as the training set, and adopt MBPP~\cite{mbpp} and HumanEval~\cite{humaneval} as test sets. 
For planning tasks, we train and evaluate on Countdown~\cite{countdown} and Sudoku~\cite{sudoku}, adopting the same training and test splits as d1~\cite{d1}.

\begin{table*}[!t]
\centering
\resizebox{\textwidth}{!}{
    \begin{tabular}{lcccccc}
    \toprule
    \multirow{2}{*}{\textbf{Model}} & \multicolumn{2}{c}{\textbf{Mathematics}} & \multicolumn{2}{c}{\textbf{Coding}} & \multicolumn{2}{c}{\textbf{Planning}} \\
    \cmidrule(lr){2-3} \cmidrule(lr){4-5} \cmidrule(lr){6-7}
                                    & \textbf{MATH500}             & \textbf{GSM8K}              & \textbf{HumanEval}        & \textbf{MBPP}             & \textbf{Sudoku}            & \textbf{Countdown}         \\ \midrule
    \multicolumn{7}{l}{\textit{Prior works with LLaDA}}                                                                                                      \\ \midrule
    d1-LLaDA~\cite{d1}                        & 40.2 & 82.1 & - & - & 16.7 & 32.0 \\
    wd1~\cite{wd1}                             & 39.0 & 82.3 & - & - & 25.2 & 46.1 \\
    LLaDA-IGPO~\cite{igpo}                     & 42.8 & 83.6 & - & - & - & - \\
    LLaDA-1.5~\cite{llada1.5}                  & 42.6 & 83.3 & 45.0* & 40.0* & - & - \\ \midrule
    \multicolumn{7}{l}{\textit{RL from LLaDA-8B-Instruct}}                                                                                                   \\ \midrule
    LLaDA-8B-Instruct~\cite{llada1.0}          & 39.6 & 79.3 & 45.1 & 39.1 & 12.0 & 19.5 \\
    + diffu-GRPO~\cite{d1}                     & 43.1 (\textcolor{green!60!black}{+3.5}) & 82.1 (\textcolor{green!60!black}{+2.8}) & 47.0 (\textcolor{green!60!black}{+1.9}) & 40.3 (\textcolor{green!60!black}{+1.2}) & 26.7 (\textcolor{green!60!black}{+14.7}) & 53.1 (\textcolor{green!60!black}{+33.6}) \\
    + VRPO-OL~\cite{llada1.5}                  & 44.1 (\textcolor{green!60!black}{+4.5}) & 83.3 (\textcolor{green!60!black}{+4.0}) & 44.8 (\textcolor{red!60!black}{-0.3}) & 41.5 (\textcolor{green!60!black}{+2.4}) & 26.1 (\textcolor{green!60!black}{+14.1}) & 84.8 (\textcolor{green!60!black}{+65.3}) \\
    + BGPO (off-policy)                        & 44.9 (\textcolor{green!60!black}{+5.3}) & \textbf{84.5} (\textcolor{green!60!black}{+5.2}) & 47.4 (\textcolor{green!60!black}{+2.3}) & 41.0 (\textcolor{green!60!black}{+1.9}) & 26.0 (\textcolor{green!60!black}{+14.0}) & 84.8 (\textcolor{green!60!black}{+65.3}) \\
    + BGPO                               & \textbf{45.7} (\textcolor{green!60!black}{+6.1}) & 84.3 (\textcolor{green!60!black}{+5.0}) & \textbf{47.6} (\textcolor{green!60!black}{+2.5}) & \textbf{41.7} (\textcolor{green!60!black}{+2.6}) & \textbf{26.9} (\textcolor{green!60!black}{+14.9}) & \textbf{87.5} (\textcolor{green!60!black}{+68.0}) \\
    \bottomrule
    \end{tabular}
}
\caption{Performance comparison between BGPO and different baselines on mathematics, coding, and planning tasks. "*" indicates we re-evaluated the model using the same code environment. The delta scores in parentheses indicate the \textcolor{green!60!black}{improvement} or \textcolor{red!60!black}{decline} compared to LLaDA-8B-Instruct.}
\label{tab:main_results}
\end{table*}

\begin{figure*}[!t]
    \centering
    \includegraphics[width=\linewidth]{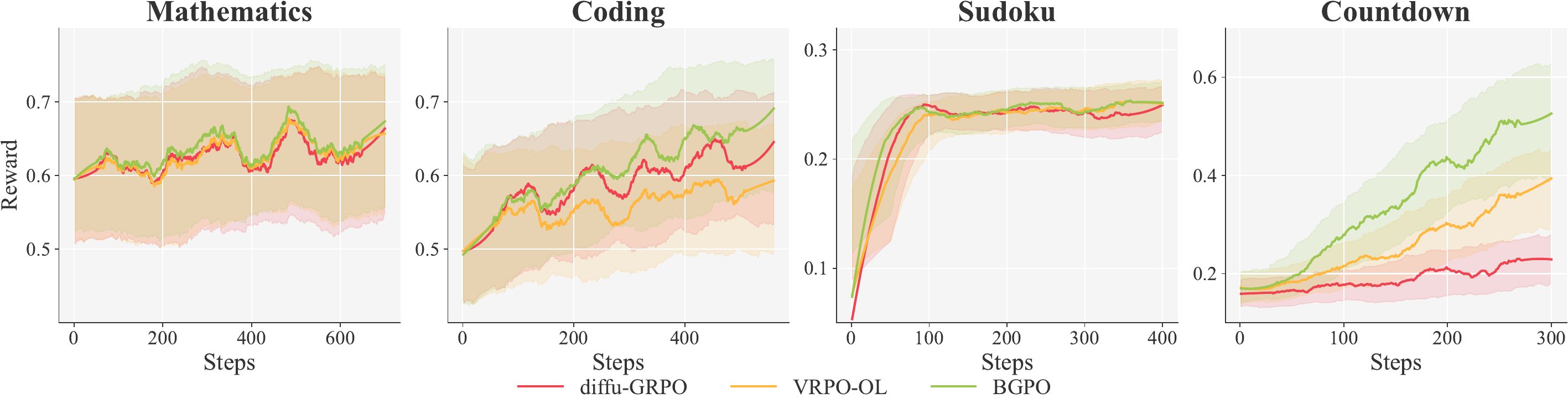}
    \caption{Training reward dynamics of diffu-GRPO, VRPO-OL and BGPO across different tasks.}
    \label{fig:reward_summary}
\end{figure*}

\paragraph{Implementation Details.} We build BGPO based on the VeRL~\cite{verl} framework.
The maximum response lengths for math problem solving, coding generation, and planning tasks are set to 512, 512, and 256, respectively. Both on-policy and off-policy settings are evaluated, where the mini-batch sizes are set to 16 and 8, respectively, with the same batch size of 16. That is, in the off-policy setting, each batch of rollout data is divided into two mini-batches for two gradient updates.
The rollout group size $G$ and the learning rate are set to 8 and $5 \times 10^{-7}$, respectively. The MC sample size $n_t$ is set to 32 for Sudoku and 16 for other tasks. See Table~\ref{tab:settings} for more detailed hyperparameters. Following~\citet{d1}, we evaluate trained models (including baselines) every 20 steps and report results from the best-performing checkpoint. All experiments are conducted on 8 $\times$H800 GPUs. 

\paragraph{Baselines.} We mainly compare BGPO with two representative RL algorithms for dLLMs that are introduced in Section~\ref{sec:preliminary}:
(1) diffu-GRPO~\cite{d1}, an on-policy algorithm that approximates the log-likelihoods with single-pass mean-field estimation;  
(2) VRPO-OL, the online version of VRPO~\cite{llada1.5} that adopts ELBO-based likelihood approximation and uses the objective in Eq~\ref{eq:vrpo_obj}. We set the MC sampling sizes of VRPO-OL to the maximum that H800 can support, i.e., $n_t=4$ for math and planning tasks and $n_t=2$ for code generation, since the prompts of coding tasks are longer. 
Besides, we also present the results of several prior works as references, including d1~\cite{d1}, wd1~\cite{wd1}, LLaDA-IGPO~\cite{igpo}, and LLaDA 1.5~\cite{llada1.5}, although their training settings are partially different from ours. 

\begin{figure*}[!t]
    \centering
    \begin{subfigure}{0.31\linewidth}
        \centering
        \includegraphics[width=\linewidth]{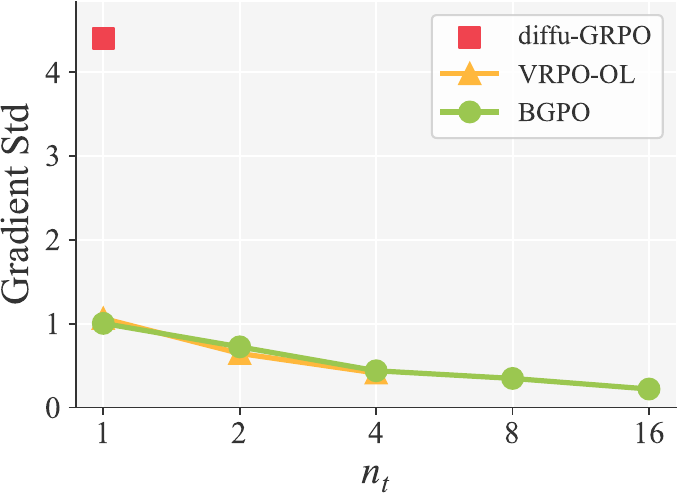}
        \label{fig:std}
    \end{subfigure}
    \begin{subfigure}{0.31\linewidth}
        \centering
        \includegraphics[width=\linewidth]{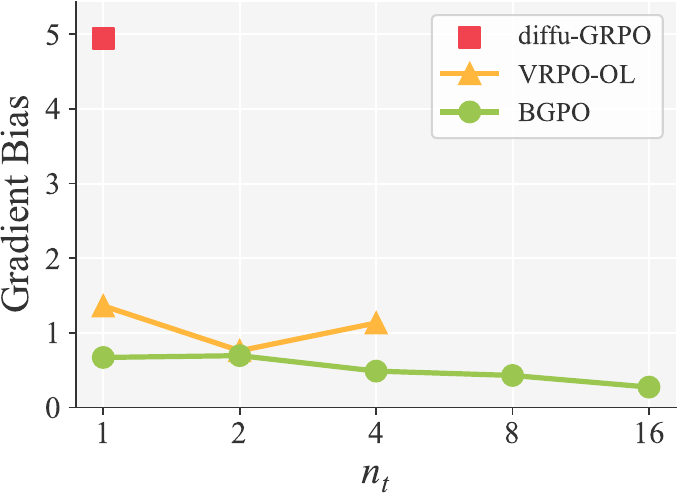}
        \label{fig:bias}
    \end{subfigure}
    \begin{subfigure}{0.32\linewidth}
        \centering
        \includegraphics[width=\linewidth]{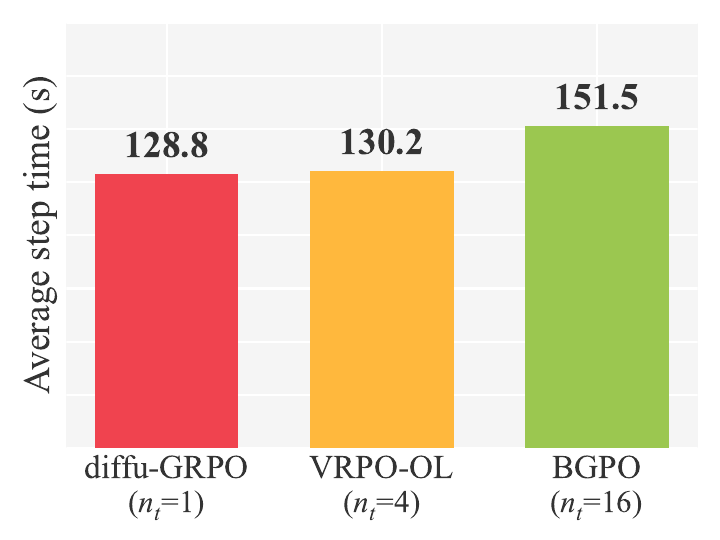}
        \label{fig:timing_step_bar}
    \end{subfigure}
    \vspace{-0.5cm}
    \caption{Left and middle: Standard deviation (std) and bias of gradients with different MC sampling size $n_t$. Right: Training speed comparison between baselines.}
    \label{fig:grad_time}
\end{figure*}

\subsection{Main Results}
Table~\ref{tab:main_results} presents the performance of BGPO and different baselines on math problem solving, code generation, and planning tasks. As shown, our BGPO algorithm achieves significant improvement over LLaDA-8B-Instruct, and also outperforms previous RL algorithms (diffu-GRPO and VRPO-OL) on all tasks, indicating that BGPO can produce a more accurate approximation of the RL objective compared to these baselines. Specifically, BGPO improves the performance of LLaDA-8B-Instruct by about 5.5\% and 2.5\% on mathematical and coding tasks, respectively, and dramatically improves the performance on Sudoku and Countdown by 14.9\% and 68.0\%. In the off-policy setting, BGPO still demonstrates strong performance, surpassing all baselines on both the math and Countdown tasks and achieving comparable improvements on other tasks. Moreover, the model trained with BGPO also outperforms all previous LLaDA-based models (e.g., wd1 and LLaDA-1.5), achieving state-of-the-art results. 

Figure~\ref{fig:reward_summary} shows the reward dynamics of BGPO, diffu-GRPO, and VRPO-OL during training on different tasks. The reward of BGPO is higher than the other two baselines in most steps. Particularly, BGPO exhibits a notably faster reward increase and higher reward on the Countdown task, where the exploration space is relatively simple. These phenomena demonstrate that the larger MC sample size of BGPO brings a more accurate optimization direction. 

\subsection{Effect of Increasing MC Sample Sizes}
To demonstrate the effect of increasing the MC sample size $n_t$ in approximating the RL objective, we train LLaDA-8B-Instruct on math problem solving using BGPO with different $n_t$. As shown in Table~\ref{tab:abla_1}, the model performance consistently improves as $n_t$ increases from 1 to 16, implying that larger MC sample sizes can produce more approximations of the RL objective. 
\begin{table}[h]
\centering
\resizebox{0.37\textwidth}{!}{
\begin{tabular}{l|cc}
\toprule
\textbf{Model}    & \textbf{MATH500} & \textbf{GSM8K} \\ \midrule
LLaDA-8B-Instruct & 39.6             & 79.3           \\
+ BGPO ($n_t=1$)  & 43.5             & 83.5           \\
+ BGPO ($n_t=2$)  & 44.1             & 82.5           \\
+ BGPO ($n_t=4$)  & 44.1             & 83.0           \\
+ BGPO ($n_t=8$)  & 45.3             & 83.9           \\
+ BGPO ($n_t=16$) & \textbf{45.7}    & \textbf{84.3}  \\ 
\bottomrule
\end{tabular}
}
\caption{Performance of BGPO with different Monte Carlo sampling size $n_t$ on mathematics benchmarks.}
\label{tab:abla_1}
\end{table}

To further illustrate this, we compare the standard deviation and bias of the loss gradients of different RL algorithms with different $n_t$ \footnote{We do not directly compare the variance and bias of the loss since the value of loss is always 0 in on-policy training.}. Specifically, we compute the gradient of a batch 8 times with different MC sample sizes, and then calculate the standard deviation for each parameter. For the bias calculation, we use the gradient with $n_t=256$ to simulate the golden gradient. As shown in the left and middle of Figure~\ref{fig:grad_time}, the gradient variance and bias of diffu-GRPO are quite large, since it adopts single-pass estimation and also partially masks the prompt. In contrast, the gradient variance and bias of VRPO-OL and BGPO gradually decrease as the MC sample size $n_t$ increases. BGPO, in particular, achieves smaller variance and bias by using a larger $n_t$, as its memory overhead remains constant regardless of $n_t$, while VRPO-OL exceeds the memory limit of H100 at $n_t=8$ (see the left of Figure~\ref{fig:intro}). This allows BGPO to have a more accurate optimization direction and more stable training, resulting in better model performance. 

\subsection{Ablation of Lemma \ref{lemma:taylor} and Lemma~\ref{lemma:jensen}}
To demonstrate the necessity of using both lemmas, we study the results on math tasks when only one of them is applied. Since the equivalence always holds in on-policy settings and thus cannot reveal the individual roles of the two lemmas as boundaries, we adopt an off-policy setting (mini-batch size of 8 with batch size of 16) instead. As shown in Table~\ref{tab:abla_lemma}, using either Lemma~\ref{lemma:taylor} or Lemma~\ref{lemma:jensen} alone leads to significant performance degradation.
\begin{table}[h]
\centering
\resizebox{0.4\textwidth}{!}{
\begin{tabular}{l|cc}
\toprule
\textbf{Model}              & \textbf{MATH500} & \textbf{GSM8K} \\ \midrule
LLaDA-8B-Instruct           & 39.6             & 79.3           \\
+ BGPO (Lemma~\ref{lemma:taylor})            & 42.1             & 83.5           \\
+ BGPO (Lemma~\ref{lemma:jensen})            & 43.1             & 83.9             \\
+ BGPO (Lemma~\ref{lemma:taylor} \& \ref{lemma:jensen})       & \textbf{44.9}    & \textbf{84.5}  \\
\bottomrule
\end{tabular}
}
\caption{Ablation study of BGPO with different theoretical components on mathematics benchmarks.}
\vspace{-0.5cm}
\label{tab:abla_lemma}
\end{table}

\begin{table*}[!t]
    \centering
    \resizebox{0.9\textwidth}{!}{
        \begin{tabular}{lcccccc}
        \toprule
        \multicolumn{1}{c}{}                                 & \multicolumn{2}{c}{\textbf{Mathematics}}                  & \multicolumn{2}{c}{\textbf{Coding}}                       & \multicolumn{2}{c}{\textbf{Planning}} \\
        \cmidrule(lr){2-3} \cmidrule(lr){4-5} \cmidrule(lr){6-7}
        \multicolumn{1}{c}{\multirow{-2}{*}{\textbf{Model}}} & \textbf{MATH500}            & \textbf{GSM8K}              & \textbf{HumanEval}          & \textbf{MBPP}               & \textbf{Sudoku}  & \textbf{Countdown} \\ \midrule
        LLaDA-8B-Instruct                                    & 39.6                        & 79.3                        & 45.1                        & 39.1                        & 6.3              & 14.5               \\
        +BGPO (train on math tasks)                          & {\color[HTML]{939393} 45.7} & {\color[HTML]{939393} 84.3} & 44.2                        & 38.6                        & 8.6              & 21.1               \\
        +BGPO (train on coding tasks)                        & 40.8                        & 80.4                        & {\color[HTML]{939393} 47.6} & {\color[HTML]{939393} 41.7} & 9.2              & 21.5               \\
        \bottomrule
        \end{tabular}
    }
    \caption{Out-of-domain performance of BGPO. The in-domain results are in gray.}
    \label{tab:ood_results}
    \end{table*}
    
\subsection{Out-of-domain Performance}
To evaluate the out-of-domain generalization capability of BGPO, we train models on math and coding tasks, respectively, and evaluate them on other tasks. As presented in Table~\ref{tab:ood_results}, the model trained on math tasks improves its performance on the planning tasks, and the model trained on coding tasks achieves improvement on both math and planning tasks, demonstrating the good generalizability of BGPO. 

\subsection{Training Efficiency Comparison}
A potential concern for BGPO is that the large MC sample size may slow each RL step and reduce training efficiency. To allay this concern, we compare the averaged training step time of BGPO with baseline methods on math problem solving, with the maximum response length set to 512. As shown in the right of Figure~\ref{fig:grad_time}, even though BGPO adopts a much larger MC sample size (i.e., $4\times$ of VRPO-OL), its average step time increases only slightly. This is because the runtime of each step is dominated by response rollout (sampling $G$ responses for each prompt) rather than by objective computation and policy updates.

\section{Related Work}

\subsection{Diffusion Large Language Models}
Diffusion large language models (dLLMs), which generate text through masked diffusion~\cite{austin2023structured, sahoo2024simple, shi2025simplified, ou2024your, nie2025scale}, have recently achieved significant advances, demonstrating performance comparable to similarly-sized autoregressive models. Among existing open-source dLLMs, DiffuLLaMA~\cite{diffullama}, Dream~\cite{dream7b}, and SDAR~\cite{sdar} are adapted from pre-trained autoregressive LLMs, while LLaDA~\cite{llada1.0} is trained from scratch using bidirectional attention by maximizing the ELBOs of log-likelihoods, presenting a complete process of pre-training and supervised fine-tuning of dLLMs. Moreover, several commercial dLLMs like Mercury~\cite{inception2025mercury}, Gemini Diffusion~\cite{deepmind2025gemini}, and Seed Diffusion~\cite{seed-diffusion} not only achieve leading performance in code generation but also offer significantly faster inference, demonstrating the practical viability of dLLMs and their promising alternative to autoregressive LLMs.

\subsection{Reinforcement Learning for dLLMs} 
Applying RL to dLLMs presents unique challenges compared to autoregressive models. The iterative, non-sequential generation process of dLLMs makes their likelihood functions intractable, necessitating the approximation of log-likelihoods for policy optimization. For instance, d1~\cite{d1} proposed diffu-GRPO, which approximates the log-likelihoods of dLLMs through single-pass mean-field estimation. Following wd1~\cite{wd1, igpo} and IGPO~\cite{igpo} also adopt this approximation approach. Though efficient, the single-pass estimation introduces notable bias relative to the exact likelihoods. Alternatively, VRPO~\cite{llada1.5} in LLaDA 1.5 approximates the log-likelihoods by their ELBOs, which is estimated via Monte Carlo (MC) sampling. Theoretically, this method can produce highly accurate approximations by using a large MC sample size. However, the practical sample size used in training is severely constrained by the GPU memory limit, since the computational graphs of all samples need to be retained for the gradient calculation of the non-linear function in the RL objective. While our BGPO algorithm addresses this memory-inefficiency limitation and supports large MC sample sizes, thereby effectively reducing the bias and variance of approximations and achieving better performance.

\section{Conclusion}
In this work, we propose BGPO, a memory-efficient RL algorithm for dLLMs that supports a large Monte Carlo sample size for approximating the sequence-level log-likelihoods and the final objective, thereby effectively reducing the bias and variance of approximations and leading to better model performance. We theoretically prove the equivalence of our BGPO objective and the previous ELBO-based objective, and conduct extensive experiments to validate the efficacy of BGPO. We hope that our work lays a solid foundation for future research on RL of dLLMs.

\section{Limitations}
In this work, we only conduct experiments on 8B-level models, since there are no larger open-source dLLMs, and our computational resources are also limited. Nonetheless, we believe our BGPO algorithm can be well applied to larger dLLMs due to its solid theoretical foundation.

\section{Ethical Considerations}
All the models and datasets used in this work are publicly published with permissible licenses. 

\section{Acknowledgement}
This work is supported by National Natural Science Foundation of China (62476150) and a grant from the Institute for Guo Qiang, Tsinghua University (2019GQB0003).

\bibliography{custom}

\appendix
\newpage
\section{Detailed Hyperparameters}
\begin{table*}[!t]
\centering
\resizebox{\linewidth}{!}{
    \setlength{\tabcolsep}{6pt}
    \begin{tabular}{lccccccc}
    \toprule
    \multirow{2}{*}{\textbf{Task}}  
    & \multirow{2}{*}{\textbf{Response length}} 
    & \multirow{2}{*}{\textbf{Diffusion step}} 
    & \multirow{2}{*}{\textbf{Block size}} 
    & \multicolumn{3}{c}{\textbf{MC sample size $n_t$}} \\
    \cmidrule(lr){5-7}
        & & & & \textbf{diffu-GRPO} & \textbf{VRPO-OL} & \textbf{BGPO} \\
    \midrule
    Mathematics & 512 / 512* & 256 / 512* & 32 / 32* & 1 & 4 & 16 \\
    Coding        & 512 / 512* & 512 / 512* & 32 / 32* & 1 & 2 & 16 \\
    Sudoku      & 256 / 256* & 128 / 256* & 32 / 32* & 1 & 4 & 32  \\
    Countdown   & 256 / 256* & 128 / 256* & 32 / 32* & 1 & 4 & 16 \\
    \bottomrule
    \end{tabular}
}
\caption{Detailed hyperparameters for different tasks. "*" denotes the different hyperparameters used in evaluation.}
\label{tab:settings}
\end{table*}

We present detailed hyperparameters of BGPO on different tasks in Table~\ref{tab:settings}. Following previous works, we adopt a block-wise decoding strategy in both training and evaluation. The choices of response length, diffusion step, and block size also follow~\citet {llada1.5} and ~\citet{d1} for obtaining the best performance. 

\section{Length Extrapolation Analysis}
\begin{table*}[!t]
    \centering
    \resizebox{0.8\linewidth}{!}{
        \begin{tabular}{lcccccc}
        \toprule
        \multicolumn{1}{c}{}                                 & \multicolumn{2}{c}{\textbf{Mathematics}}                  & \multicolumn{2}{c}{\textbf{Coding}}                       & \multicolumn{2}{c}{\textbf{Planning}} \\
        \cmidrule(lr){2-3} \cmidrule(lr){4-5} \cmidrule(lr){6-7}
        \multicolumn{1}{c}{\multirow{-2}{*}{\textbf{Length}}} & \textbf{MATH500}            & \textbf{GSM8K}              & \textbf{HumanEval}          & \textbf{MBPP}               & \textbf{Sudoku}  & \textbf{Countdown} \\ \midrule
        256    & 40.8 & 76.4 & 40.9 & \textbf{42.7} & \textbf{26.9} & \textbf{87.5} \\
        512    & \textbf{45.7} & \textbf{84.3} & 47.6 & 41.7 & 24.6 & 84.4 \\
        1024   & 41.9 & 83.8 & \textbf{48.9} & 41.1 & 25.9 & 84.0 \\
        2048   & 44.1 & 83.7 & 48.6 & 41.3 & 23.5 & 75.4 \\
        \bottomrule
        \end{tabular}
    }
    \caption{Inference-time length scaling results across different benchmarks.}
    \label{tab:length_extrapolation}
\end{table*}
Table~\ref{tab:length_extrapolation} presents our inference-time length scaling results across multiple benchmarks.
As shown, performance generally peaks when the inference length matches the training length, but deteriorates as the sequence length increases. This decline can be attributed to the limited long-chain-of-thought reasoning capabilities of current open-source dLLMs. As the sequence length grows, dLLMs struggle to retain and process information over extended contexts, resulting in diminished performance. This observation is consistent with prior works \cite{llada1.5,d1}.

\end{document}